\def\year{2020}\relax
\documentclass[letterpaper]{article} % DO NOT CHANGE THIS
\usepackage{aaai20}  % DO NOT CHANGE THIS
\usepackage{times}  % DO NOT CHANGE THIS
\usepackage{helvet} % DO NOT CHANGE THIS
\usepackage{courier}  % DO NOT CHANGE THIS
\usepackage[hyphens]{url}  % DO NOT CHANGE THIS
\usepackage{graphicx} % DO NOT CHANGE THIS
\usepackage{graphicx}  % DO NOT CHANGE THIS
\newcommand{\citet}[1]{\citeauthor{#1} \shortcite{#1}}
\newcommand{\citep}{\cite}

\def\eg{\emph{e.g.}}
\def\ie{\emph{i.e.}}
\def\etc{\emph{etc.}}
\def\etal{\emph{et al.}}
\usepackage{multirow}
\usepackage[tight,normalsize,sf,SF]{subfigure}
\usepackage{array} 
\usepackage{amsmath}

\title{Learning Transferable Adversarial Examples via Ghost Networks}
\author{
Yingwei Li\textsuperscript{\rm 1}~~~~
Song Bai\textsuperscript{\rm 2}~~~~
Yuyin Zhou\textsuperscript{\rm 1}~~~~
Cihang Xie\textsuperscript{\rm 1}~~~~
Zhishuai Zhang\textsuperscript{\rm 1}~~~~ 
Alan Yuille\textsuperscript{\rm 1} \\
\textsuperscript{\rm 1}Johns Hopkins University \qquad\qquad \textsuperscript{\rm 2}University of Oxford \\
{\normalsize yingwei.li@jhu.edu, \{songbai.site, zhouyuyiner, cihangxie306, zhshuai.zhang, alan.l.yuille\}@gmail.com}
}
 
\begin{document}

\maketitle

%%%%%%%%% ABSTRACT
\begin{abstract}
Recent development of adversarial attacks has proven that ensemble-based methods outperform traditional, non-ensemble ones in black-box attack. However, as it is computationally prohibitive to acquire a family of diverse models, these methods achieve inferior performance constrained by the limited number of models to be ensembled.
%these methods generally require a family of diverse models, and ensembling them together afterward, both of which are computationally expensive. 

In this paper, we propose Ghost Networks to improve the transferability of adversarial examples. The critical principle of ghost networks is to apply feature-level perturbations to an existing model to potentially create a huge set of diverse models. After that, models are subsequently fused by longitudinal ensemble.
Extensive experimental results suggest that the number of networks is essential for improving the transferability of adversarial examples, but it is less necessary to independently train different networks and ensemble them in an intensive aggregation way. Instead, our work can be used as a computationally cheap and easily applied plug-in to improve adversarial approaches both in single-model and multi-model attack, compatible with residual and non-residual networks. By reproducing the NeurIPS 2017 adversarial competition, our method outperforms the No.1 attack submission by a large margin, demonstrating its effectiveness and efficiency. Code is available at \url{https://github.com/LiYingwei/ghost-network}.

%Compared to traditional ensemble methods, both steps require almost no extra time and space consumption.

%In this paper, we propose Ghost Networks to generate transferable adversarial examples efficiently. The critical principle of ghost networks is to apply feature-level perturbations to an existing model to potentially create a huge set of diverse models. After that, models are subsequently fused by longitudinal ensemble. Compared to traditional ensemble methods, both steps require almost no extra time and space consumption. 

%Extensive experimental results suggest that the number of networks is essential for improving the transferability of adversarial examples, but it is less necessary to independently train different networks and ensemble them in an intensive aggregation way. Instead, our work can be used as a computationally cheap and easily applied plug-in to improve adversarial approaches both in single-model and multi-model attack, compatible with residual and non-residual networks. By reproducing the NeurIPS 2017 adversarial competition, our method outperforms the No.1 attack submission by a large margin, demonstrating its effectiveness and efficiency.
\end{abstract}

\section{Introduction} \label{sec:intro}
In recent years, Convolutional Neural Networks (CNNs) have greatly advanced performance in various vision tasks, including image recognition~\cite{he2016identity,krizhevsky2012imagenet,simonyan2015very}, object detection~\cite{Ren_2015_Faster,Girshick_2015_Fast}, and semantic segmentation~\cite{Chen_2016_DeepLab},~\etc. However, it has been observed~\cite{szegedy2013intriguing,goodfellow2014explaining} that adding human imperceptible perturbations to input image can cause CNNs to make incorrect predictions even if the original image can be correctly classified. These intentionally generated images are usually called adversarial examples~\cite{goodfellow2014explaining,kurakin2016adversarial,szegedy2013intriguing}. Besides image classification, adversarial examples also exist on other tasks~\cite{xie2017adversarial,bai2019adversarial,sun2019node,tang2019robust,li2019volumetric}.

% In recent years, Convolutional Neural Networks (CNNs) have advanced performance in various vision tasks. However, it has been observed that attacking CNNs by adding human imperceptible perturbations to input images can cause networks to make incorrect predictions~\cite{szegedy2013intriguing}. These perturbed images are termed as adversarial examples. 
% \textcolor{blue}{Many works are paying attention to how to learn adversarial examples to attack CNNs~\cite{carlini2016towards,kurakin2016adversarial,papernot2016transferability,dong2017boosting}.}

%\cite{carlini2016towards} and I-FGSM~\cite{kurakin2016adversarial} are later developed to attack a white-box model (attackers can access the model).
%\footnote{It is strange to say xxx et al and a method are developed...}
% Defense techniques based on randomization~\cite{xie2017mitigating,Liu_2018_ECCV}, input transformation~\cite{guo2017countering}, and training~\cite{tramer2017ensemble,madry2017towards,liao2018defense} are also studied.
\begin{figure}[tb]
\begin{center}
\includegraphics[width=0.95\linewidth]{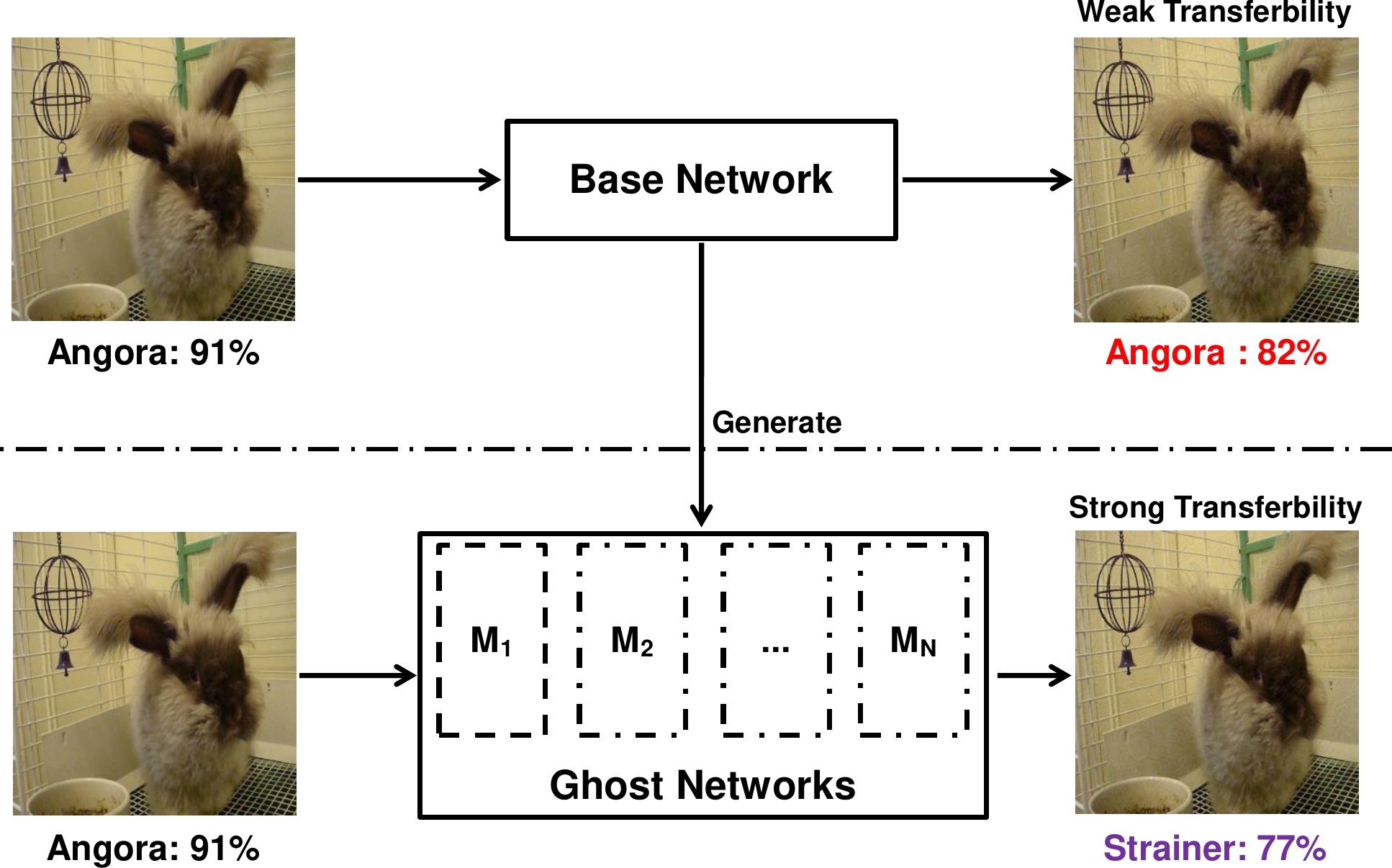}
\end{center}
% \vspace{-2ex}
\caption{An illustration of the capacity of the proposed ghost networks in learning transferable adversarial examples. The base model is ResNet-50, which is used to generate adversarial examples and to generate ghost networks. The evaluation is done on Inception v3.}
\label{fig:ill_trans}
% \vspace{-3ex}
\end{figure}

Two attack settings are later developed, \ie, white-box attack and black-box attack.
In white-box attack, attackers can access the model~\cite{carlini2016towards,kurakin2016adversarial}.
By contrast, in black-box attack,
attackers cannot access the target model. A typical solution is to generate adversarial examples with strong transferability (transferable adversarial examples).

The transferability of adversarial examples refers to the property that the same input can successfully attack different models~\cite{szegedy2013intriguing}. Taking advantage of the transferability, \cite{papernot2016transferability} develop a black-box (attackers cannot access the target model) attack system by attacking a substitute model. \cite{liu2016delving} suggest attacking an ensemble of substitute models could improve transferability. Based on \cite{liu2016delving} and I-FGSM, several methods are developed to further improve the transferability by smoothing gradient~\cite{dong2017boosting,Zhou_2018_ECCV}. In this work, we focus on learning transferable adversarial examples for black-box attack.

% Meanwhile, defense techniques based on randomization~\cite{xie2017mitigating,Liu_2018_ECCV}, input transformation~\cite{guo2017countering}, and training~\cite{tramer2017ensemble,madry2017towards} are also studied.

Focusing on the transferability, many attempts have been made, such as attacking a substitute model~\cite{papernot2016transferability} or an ensemble of multiple substitute models~\cite{liu2016delving,dong2017boosting,Zhou_2018_ECCV}.
In particular, the ensemble-based attacks obtain much better performance than the non-ensemble ones, and thus have attracted many attentions.
Almost all top-ranked entries in competitions use ensemble-based attacks~\cite{kurakin2018adversarial}.

% The transferability of adversarial examples refers to the property that the same input can successfully attack different models~\cite{szegedy2013intriguing}. Taking advantage of the transferability, \cite{papernot2016transferability} develop a black-box (attackers cannot access the target model) attack system by attacking a substitute model. \cite{liu2016delving} suggest attacking an ensemble of substitute models could improve transferability. Based on \cite{liu2016delving} and I-FGSM, several methods are developed to further improve the transferability by smoothing gradient~\cite{dong2017boosting,Zhou_2018_ECCV}.

However, the ensemble-based attacks suffer from expensive computational overhead, making it difficult to generate transferable adversarial examples efficiently. First, in order to acquire good (\ie,~low test error) and diverse (\ie,~converge at different local minima) models, people usually independently train them from scratch. Second, to leverage their complementarity, existing methods adopt an intensive aggregation way to fuse the outputs of those networks (\eg,~logits). Consequently, attacking methods in competitions (like \citet{kurakin2018adversarial}) generally ensemble at most only ten networks restricted by the high computational cost. However, efficiently attacking a huge ensemble of models is critical.

How to improve the transferability of adversarial examples without additional cost remains a challenging task. \citet{baluja2017adversarial,poursaeed2017generative,xiao2018generating} suggest that re-training networks can achieve high transferability. \citet{chen2017zoo,bhagoji2018practical,guo2018low} propose query-based methods to attack black-box model without substitute models, which require extensive information from the target model. In conclusion, acquiring and integrating information from various models to approximate the target model is the key to achieving better transferability. However, most works are inefficient and inadequate to learn adversarial examples with strong transferability. Our work addresses this issue.

In this paper, we propose a highly efficient alternative called Ghost Networks to address this issue. As shown in Fig.~\ref{fig:ill_trans}, the basic principle is to generate a vast number of virtual models built on a base network (a network trained from scratch). The word ``virtual" means that these networks are not stored or trained (therefore termed as ghost networks). Instead, they are generated by imposing erosion on certain intermediate structures of the base network on-the-fly. However, with an increasing number of models we have, a standard ensemble~\cite{liu2016delving} would be problematic owing to its complexity. Accordingly, we propose Longitudinal Ensemble, a specific fusion method for ghost networks, which conducts an implicit ensemble during attack iterations. Consequently, adversarial examples can be easily generated without sacrificing computational efficiency.

To summarize, the contributions of our work are divided into three folds: 1) Our work is the first one to explore network erosion to learn transferable adversarial examples, not solely relying on multi-network ensemble. 2) We observe that the number of different networks used for ensemble (intrinsic networks) is essential for transferability. However, it is less necessary to train different models independently. Ghost networks can be a competitive alternative with extremely low complexity. 3) Ghost network is generic. Seemingly an ensemble method for multi-model attacks, it can also be applied to single-model attacks where only one trained model is accessible. Furthermore, it is also compatible with various network structures, attack methods, and adversarial settings.

Extensive experimental results demonstrate our method improves the transferability of adversarial examples, acting as a computationally cheap plug-in. In particular, by reproducing NeurIPS 2017 adversarial competition~\cite{kurakin2018adversarial}, our work outperforms the No.1 attack submission by a large margin, demonstrating its effectiveness and efficiency.

\section{Backgrounds} \label{sec:related work}
This section introduces two iterative-based methods, Iterative Fast Gradient Sign Method (I-FGSM)~\cite{kurakin2016adversarial} and Momentum I-FGSM (MI-FGSM)~\cite{dong2017boosting}.

% \vspace{1ex}\noindent\textbf{I-FGSM} initializes an adversarial example $I_{0}^{\text{adv}} = I$ and then iteratively updates it by
% \begin{equation} \label{eq:IFGSM update}
% I_{n+1}^{\text{adv}} = \text{Clip}_{I}^{\epsilon} \{I_{n}^{\text{adv}} + \alpha\text{sign}\left(\nabla_{I} L(I_{n}^{\text{adv}}, y^{\text{true}}; \theta)\right)\},
% \end{equation}
% where $L$ is the loss function of a network with parameter $\theta$. The clip function $\text{Clip}_{I}^{\epsilon}$ ensures the generated adversarial example within the $\epsilon$-ball of the original image $I$ with ground-truth $y^{\text{true}}$. $n$ is the iteration number, and $\alpha$ is the step size.

% \vspace{1ex}\noindent\textbf{MI-FGSM} integrates the momentum term into the attack process to stabilize update directions and escape from poor local maxima. At the $n^\text{th}$ iteration, the accumulated gradient $g_{n+1}$ is calculated by
% \begin{equation} \label{eq: momentum term}
% g_{n+1} = \mu \cdot g_n + \frac{\nabla_{I} L(I_{n}^{\text{adv}}, y^{\text{true}}; \theta)}{||\nabla_{I} L(I_{n}^{\text{adv}}, y^{\text{true}}; \theta)||_1},
% \end{equation}
% where $\mu$ is the momentum decay factor. The sign of $g_{n+1}$ is then used to generate the adversarial example, by
% \begin{equation}
% I_{n+1}^{\text{adv}} = \text{Clip}_{I}^{\epsilon}\{I_{n}^{\text{adv}} + \alpha\text{sign}(g_{n+1})\}.
% \end{equation}

\vspace{1ex}\noindent\textbf{I-FGSM} was proposed by Kurakin~\etal~\cite{kurakin2016adversarial}, and learns the adversarial example $I^{\text{adv}}$ by
\begin{equation} \label{eq:IFGSM update}
\begin{split}
&I_{0}^{\text{adv}} = I, \\
&I_{n+1}^{\text{adv}} = \text{Clip}_{I}^{\epsilon} \{I_{n}^{\text{adv}} + \alpha\text{sign}\left(\nabla_{I} L(I_{n}^{\text{adv}}, y^{\text{true}}; \theta)\right)\},
\end{split}
\end{equation}
where $L$ is the loss function of a network with parameter $\theta$. $\text{Clip}_{I}^{\epsilon}$ is the clip function which ensures the generated adversarial example is within the $\epsilon$-ball of the original image $I$ with ground-truth label $y^{\text{true}}$. $n$ is the iteration number and $\alpha$ is the step size.

\vspace{1ex}\noindent\textbf{MI-FGSM} was proposed by Dong~\etal~\cite{dong2017boosting}, and integrates the momentum term into the attack process to stabilize the update directions and escape from poor local maxima. At the $n$-th iteration, the accumulated gradient $g_{n+1}$ is calculated by:
\begin{equation} \label{eq: momentum term}
g_{n+1} = \mu \cdot g_n + \frac{\nabla_{I} L(I_{n}^{\text{adv}}, y^{\text{true}}; \theta)}{||\nabla_{I} L(I_{n}^{\text{adv}}, y^{\text{true}}; \theta)||_1},
\end{equation}
where $\mu$ is the decay factor of the momentum term. The sign of the accumulated gradient $g_{n+1}$ is then used to generate the adversarial example, by
\begin{equation}
I_{n+1}^{\text{adv}} = \text{Clip}_{I}^{\epsilon}\{I_{n}^{\text{adv}} + \alpha\text{sign}(g_{n+1})\}.
\end{equation}

\section{Ghost Networks} \label{sec:method}
The goal of this work is to learn transferable adversarial examples. Given a clean image $I$, we want to find an adversarial example $I^\text{adv}=I+r$, which is still visually similar to $I$ after adding adversarial noise $\|r\|_\infty<\epsilon$ but fools the classifier. %In order to improve transferability, we choose to attack multiple models simultaneously. However, unlike existing work~\cite{liu2016delving}, we propose Ghost Networks, a highly efficient algorithm that can both generate and ensemble procedure.

Without additional cost, we generate a huge number of ghost networks from a single trained model for later attack by applying feature-level perturbations to non-residual and residual based networks in the next two subsections respectively.
% These ghost networks are efficiently emsembled by our customized fusion method, logitudinal ensemble, see Sec.~\ref{sec:longitudinal}.
Then we present an efficient customized fusion method, longitudinal ensemble, leading to ensemble a huge amount of ghost networks possible.

\subsection{Dropout Erosion} \label{sec:dropout}
\vspace{1ex}\noindent\textbf{Revisit Dropout.}~Dropout~\cite{srivastava2014dropout} is one of the most popular techniques in deep learning. By randomly dropping out units from the model during training phase, dropout can prevent deep neural networks from overfitting. 
% Let $x_l$ be the activation in the $l^\text{th}$ layer, at the training time, the output $y_l$ after a dropout layer can be defined as
% \begin{equation} \label{eq:dropout}
% \begin{split}
% y_l=r_l*x_l, \quad
% r_l\sim\text{Bernoulli}(p),
% \end{split}
% \end{equation}
Let $x_l$ be the activation in the $l^\text{th}$ layer, at training time, the output $y_l$ after the dropout layer can be mathematically defined as
\begin{equation} \label{eq:dropout}
\begin{split}
& y_l=r_l*x_l, \\
& r_l\sim\text{Bernoulli}(p),
\end{split}
\end{equation}
where $*$ denotes an element-wise product and $\text{Bernoulli}(p)$ denotes the Bernoulli distribution with the probability $p$ of elements in $r_l$ being $1$. At the test time, units in $x_l$ are always present, thus to keep the output $y_l$ the same as the expected output at the training time, $y_l$ is set to be $p x_l$.

\vspace{1ex}\noindent\textbf{Perturb Dropout.}~Dropout provides an efficient way of approximately combining different neural network architectures and thereby prevents overfitting. Inspired by this, we propose to generate ghost networks by inserting the dropout layer. To make ghost networks as diverse as possible, we \textbf{densely} apply dropout to every block \textbf{throughout} the base network, \textit{rather than simply enable default dropout layers~\cite{carlini2017adversarial}}. From our preliminary experiments, the latter cannot provide transferability. Therefore, diversity is not limited to high-level features but applied to all feature levels. 

% Let $f_l$ be the function between the $i^\text{th}$ and $(i+1)^\text{th}$ layer,~\ie,~$x_{l+1}=f_l(x_l)$, then the output of $f_l$ after applying dropout erosion,~\ie,~$g_l(x_l)$, is
% \begin{equation} \label{eq:drop-core}
% \begin{split}
% g_l(x_l) = f_l\Big(\frac{r_l*x_l}{1-\Lambda}\Big), \quad
% r_l\sim\text{Bernoulli}(1-\Lambda),
% \end{split}
% \end{equation}
Let $f_l$ be the function between the $i^\text{th}$ and $(i+1)^\text{th}$ layer,~\ie,~$x_{l+1}=f_l(x_l)$, then the output $g_l(x_l)$ after applying dropout erosion is
\begin{equation} \label{eq:drop-core}
\begin{split}
& g_l(x_l) = f_l\Big(\frac{r_l*x_l}{1-\Lambda}\Big), \\
& r_l\sim\text{Bernoulli}(1-\Lambda),
\end{split}
\end{equation}
where $\Lambda=1-p$, and $p$ has the same meaning as in Eq.~\eqref{eq:dropout}, indicating the probability that $x_l$ is preserved. To keep the expected input of $f_l(\cdot)$ consistent after erosion, the activation of $x_l$ should be divided by $1-\Lambda$.

During the inference, the output feature after $(L-1)$-th dropout layer~($L>l$) is
\begin{equation} \label{eq: drop-fw}
x_{L} = g_{L-1}\circ g_{L-2}\circ g_{L-3}\circ\cdots\circ g_{l}(x_l),
\end{equation}
where $\circ$ denotes composite function,~\ie,~$g\circ f(x)=g\left(f(x)\right)$.

By combining Eq.~\eqref{eq:drop-core} and Eq.~\eqref{eq: drop-fw}, we observe that when $\Lambda=0$ (means $p = 1$), all elements in $r_l$ equal to $1$. In this case, we do not impose any perturbations to the base network. When $\Lambda$ gradually increases to $1$ ($p$ decreases to $0$), the ratio of elements dropped out is $\Lambda$. In other words, $(1-\Lambda)$ of elements can be back-propagated. Hence, larger $\Lambda$ implies a heavier erosion on the base network. Therefore, we define $\Lambda$ to be the \textit{magnitude of erosion}.

When perturbing dropout layers, the gradient in back-propagation can be written as
\begin{equation} \label{eq: drop-bp}
\frac{\partial x_{L}}{\partial x_l}=\prod_{l\leq i< L}\left(\frac{r_i}{1-\Lambda} * \frac{\partial }{\partial x_i}f_i\Big(\frac{r_i*x_i}{1-\Lambda}\Big)\right).
\end{equation}
As shown in Eq.~\eqref{eq: drop-bp}, deeper networks with larger $L$ are influenced more easily according to the product rule. We will experimentally analyze the impact of $\Lambda$ in the experiments part.

\vspace{1ex}\noindent\textbf{Generate Ghost Network.}~The generation of ghost networks via perturbing dropout layer proceeds in three steps: 1) randomly sample a parameter set from the Bernoulli distribution $r=\{r_1,r_2...,r_l,...,r_L\}$; 2) apply Eq.~\eqref{eq:drop-core} to the base network with the parameter set $r$ and get the perturbed network; 3) repeat step 1) and step 2) to independently sample $r$ for $N$ times and obtain a pool of ghost networks $M=\{M_1,M_2,...,M_N\}$ which can be used for adversarial attacks.

\begin{figure}[tb]
\centering
\includegraphics[width=0.85\linewidth]{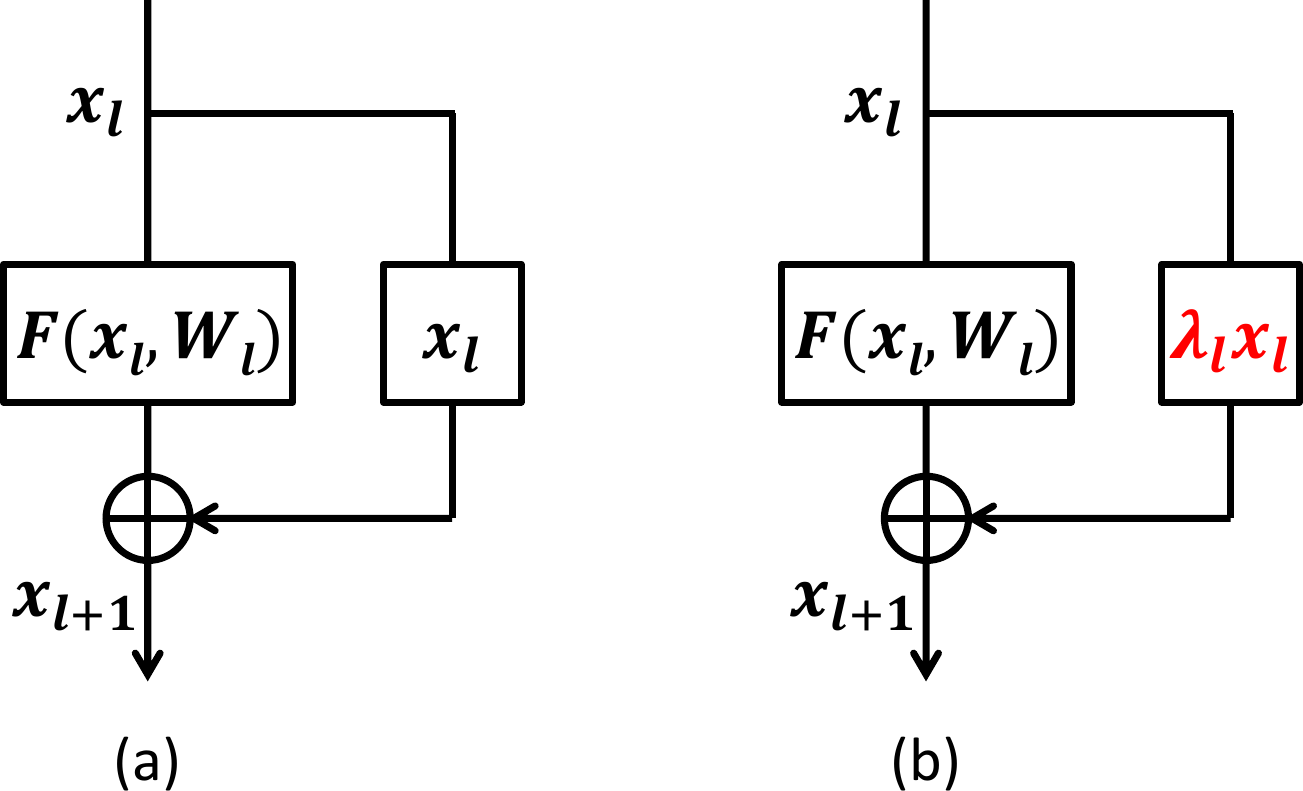}
% \includegraphics[width=0.76\linewidth]{skip_connection_cropped.pdf}
% \vspace{-2ex}
\caption{An illustration of skip connection (a, Eq.~\eqref{eq: raw-res}) and skip connection erosion (b, Eq.~\eqref{eq: res}).}
\label{fig:residual}
% \vspace{-3ex}
\end{figure}

\subsection{Skip Connection Erosion} \label{sec:residual}
\vspace{1ex}\noindent\textbf{Revisit Skip Connection.} \citet{he2016deep} propose skip connections in CNNs, which makes it feasible to train very deep neural networks. 
The residual block is defined by
\begin{equation}\label{eq: raw-res}
x_{l+1}=h(x_l)+F(x_l,W_l),
\end{equation}
where $x_l$ and $x_{l+1}$ are the input and output to the $l$-th residual block with the weights $W_l$. $F(\cdot)$ denotes the residual function. As suggested in \citet{he2016identity}, it is crucial to uses the identity skip connection,~\ie,~$h(x_l) = x_l$, to facilitate the residual learning process, otherwise the network may not converge to a good local minima.

\vspace{1ex}\noindent\textbf{Perturb Skip Connection.}~Following the principle of skip connection, we propose to perturb skip connections to generate ghost networks.

Specifically, the network weights are first learned using identity skip connections, then switched to the randomized skip connection (see Fig.~\ref{fig:residual}). To this end, we apply randomized modulating scalar $\lambda_l$ to the $l$-th residual block by
\begin{equation}\label{eq: res}
x_{l+1}=\lambda_l x_l+F(x_l,W_l),
\end{equation}
where $\lambda_l$ is drawn from the uniform distribution $U[1-\Lambda, 1+\Lambda]$. One may have observed several similar formulations on skip connection to improve the classification performance,~\eg,~the gated inference in~\citet{veit2018convolutional} and lesion study in~\citet{veit2016residual}. However, our work focuses on attacking the model with a randomized perturbation on skip connection,~\ie,~the model is not trained via Eq.~\eqref{eq: res}.

During inference, the output after $(L-1)$th layer is
\begin{equation}\label{eq: res-fw}
x_L=(\prod_{i=l}^{L-1}\lambda_i)x_l+\sum_{i=l}^{L-1}(\prod_{j=i+1}^{L-1}\lambda_j)F(x_i,W_i).
\end{equation}
The gradient in back-propagation is then written as
\begin{equation}\label{eq: res-bp}
\frac{\partial x_L}{\partial x_l} = (\prod_{i=l}^{L-1}\lambda_i)+\sum_{i=l}^{L-1} (\prod_{j=i+1}^{L-1}\lambda_j)\frac{\partial F(x_i,W_i)}{\partial{x_l}}.
\end{equation}
Similar to the analysis of dropout erosion, we conclude from Eq.~\eqref{eq: res-fw} and Eq.~\eqref{eq: res-bp} that a larger $\Lambda$ will have a greater influence on the base network and deeper networks are easily influenced.

\vspace{1ex}\noindent\textbf{Generate Ghost Network.}~The generation of ghost networks via perturbing skip connections is similar to that via perturbing the dropout layer. The only difference is we need to sample a set of modulating scalars $\lambda=\{\lambda_1,\lambda_2,...,\lambda_L\}$ from the uniform distribution for each skip connection.

\begin{figure}[tb]
\centering
\subfigure[]
{
\begin{minipage}[tb]{0.43\textwidth}
% \begin{minipage}[tb]{0.36\textwidth}
\includegraphics[width = 1\textwidth]{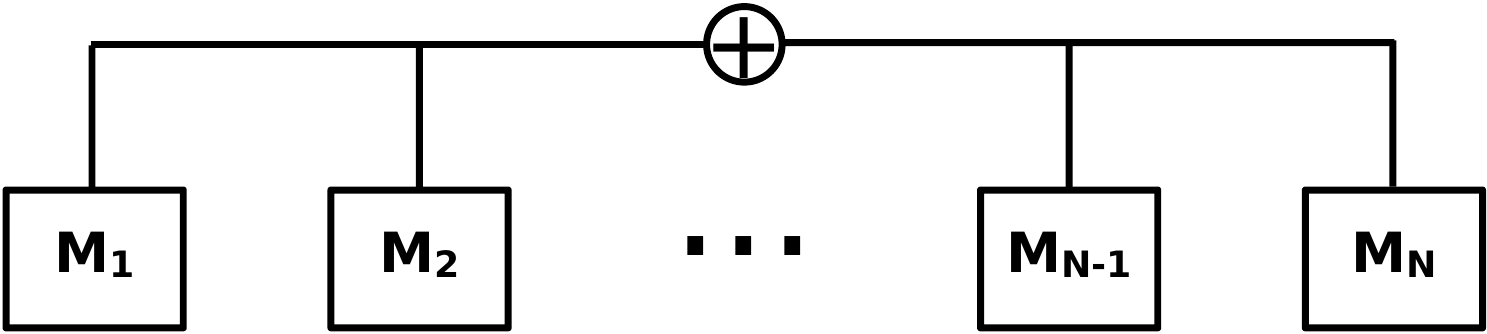}
\end{minipage}
%\label{fig:xx}
}

\subfigure[]
{
\begin{minipage}[tb]{0.43\textwidth}
% \begin{minipage}[tb]{0.36\textwidth}
\includegraphics[width = 1\textwidth]{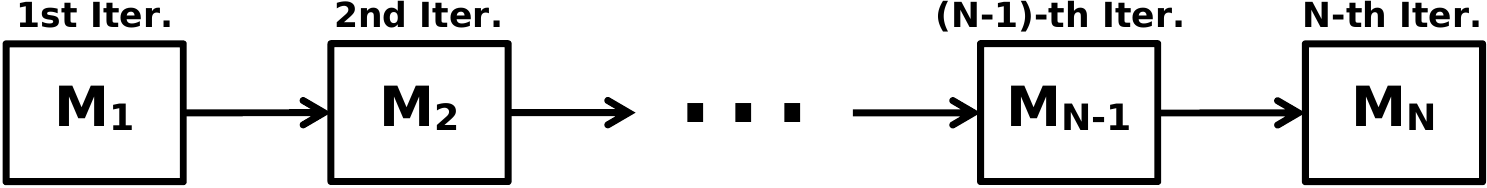}
\end{minipage}
%\label{fig:xx}
}
% \vspace{-2ex}
\caption{The illustration of the standard ensemble (a) and the proposed longitudinal ensemble (b).}
\label{fig:long_method}
% \vspace{-3ex}
\end{figure}

\subsection{Longitudinal Ensemble} \label{sec:longitudinal} 
The existing iteration-based ensemble-attack approach~\cite{liu2016delving} require averaging the outputs (\eg,~logits, classification probabilities, losses) of different networks. However, such a standard ensemble is too costly and inefficient for us because we can readily obtain a huge candidate pool of qualified neural models by using Ghost Networks.

To remedy this, we propose longitudinal ensemble, a specific fusion method for Ghost Networks, which constructs an implicit ensemble of the ghost networks by randomizing the perturbations during iterations of adversarial attack (\eg,~I-FGSM~\cite{kurakin2016adversarial} and MI-FGSM~\cite{dong2017boosting}). Suppose we have a base model $B$, which can generate a pool of networks $M=\{M_1,M_2,...,M_N\}$, where $N$ is the model number. The critical step of longitudinal ensemble is that at the $j$-th iteration, we attack the ghost network $M_j$ only. In comparison, for each iteration, standard ensemble methods require fusing gradients of all the models in the model pool $M$, leading to high computational cost. We illustrate the difference between the standard ensemble and the longitudinal ensemble method in Fig.~\ref{fig:long_method}. 

The longitudinal ensemble shares the same prior as~\citet{liu2016delving} that if an adversarial example is generated by attacking multiple networks, it is more likely to transfer to other networks. However, longitudinal ensemble method removes duplicated computations by sampling only one model from the pool rather than using all models in each iteration.

There are three noteworthy comments here. First, ghost networks are never stored or trained, reducing both additional time and space cost. Second, it is evident from Fig.~\ref{fig:long_method} that attackers can combine \citet{liu2016delving} and longitudinal ensemble of ghost networks. Finally, it is easy to extend longitudinal ensemble to multi-model attack by treating each base model as a branch (details are in experimental evaluations).

\begin{figure}[tb]
\centering
\subfigure[]
{
\begin{minipage}[t]{0.23\textwidth}
\includegraphics[width = 0.942\textwidth]{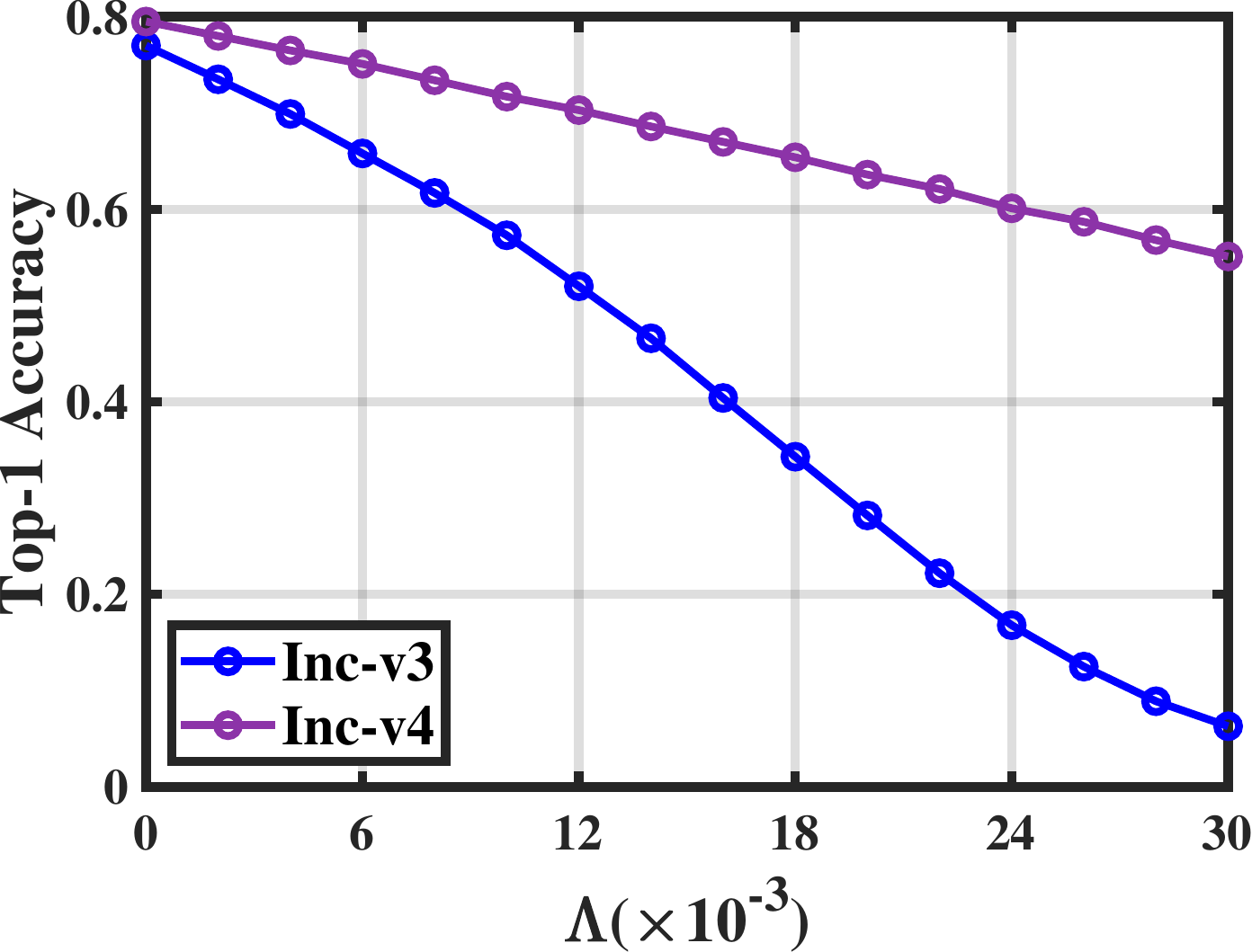}
\end{minipage}
\label{fig:dropout_acc}
}
\hspace{-2ex}
\subfigure[]
{
\begin{minipage}[t]{0.23\textwidth}
\includegraphics[width = 0.942\textwidth]{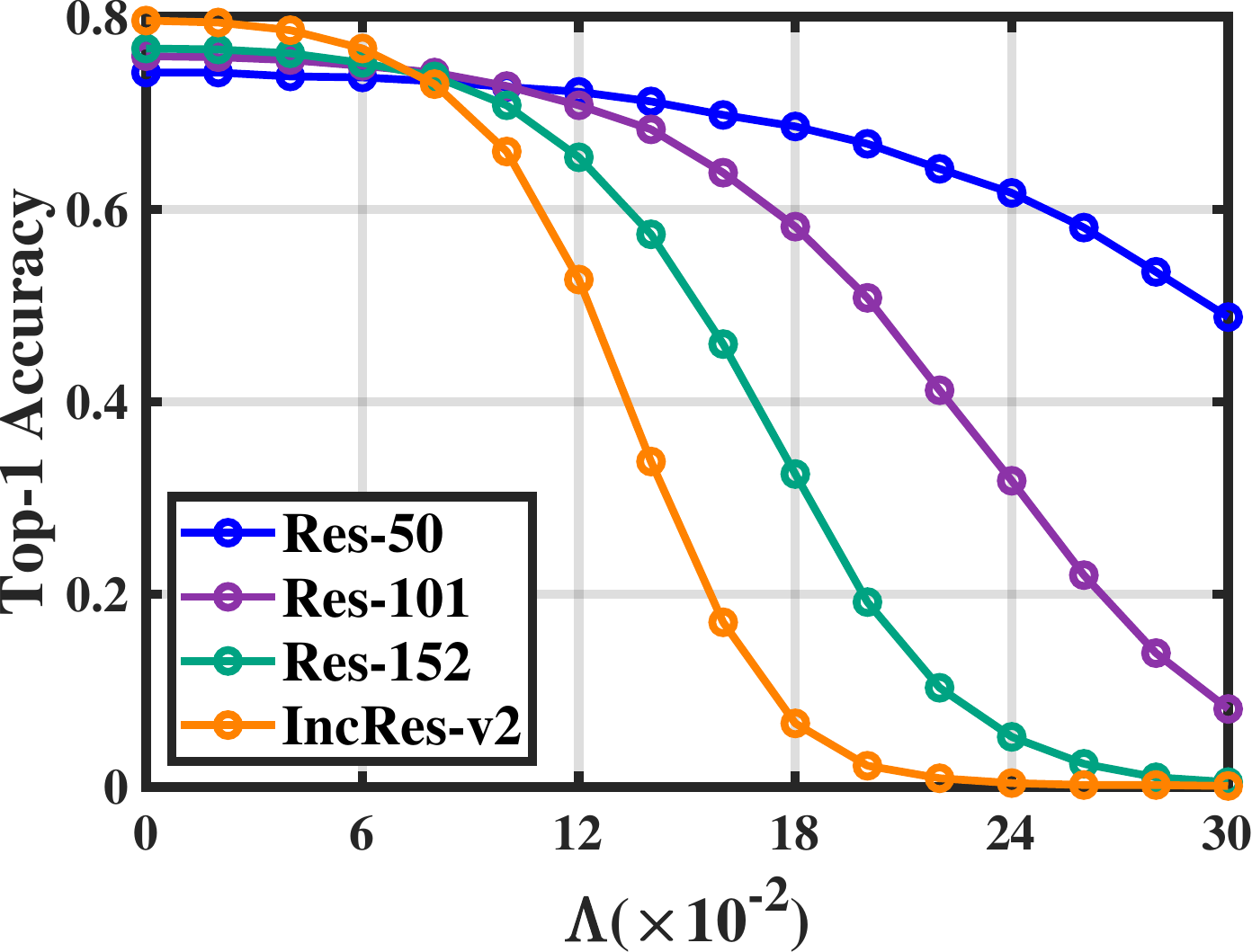}
\end{minipage}
\label{fig:residual_acc}
}
% \vspace{-2ex}
\caption{The top-1 accuracy of dropout erosion (a) and skip connection erosion (b) with different magnitude $\Lambda$ .}
\label{fig:GhostAcc}
% \vspace{-2ex}
\end{figure}

\begin{figure}[tb]
\begin{center}
\includegraphics[width=0.9\linewidth]{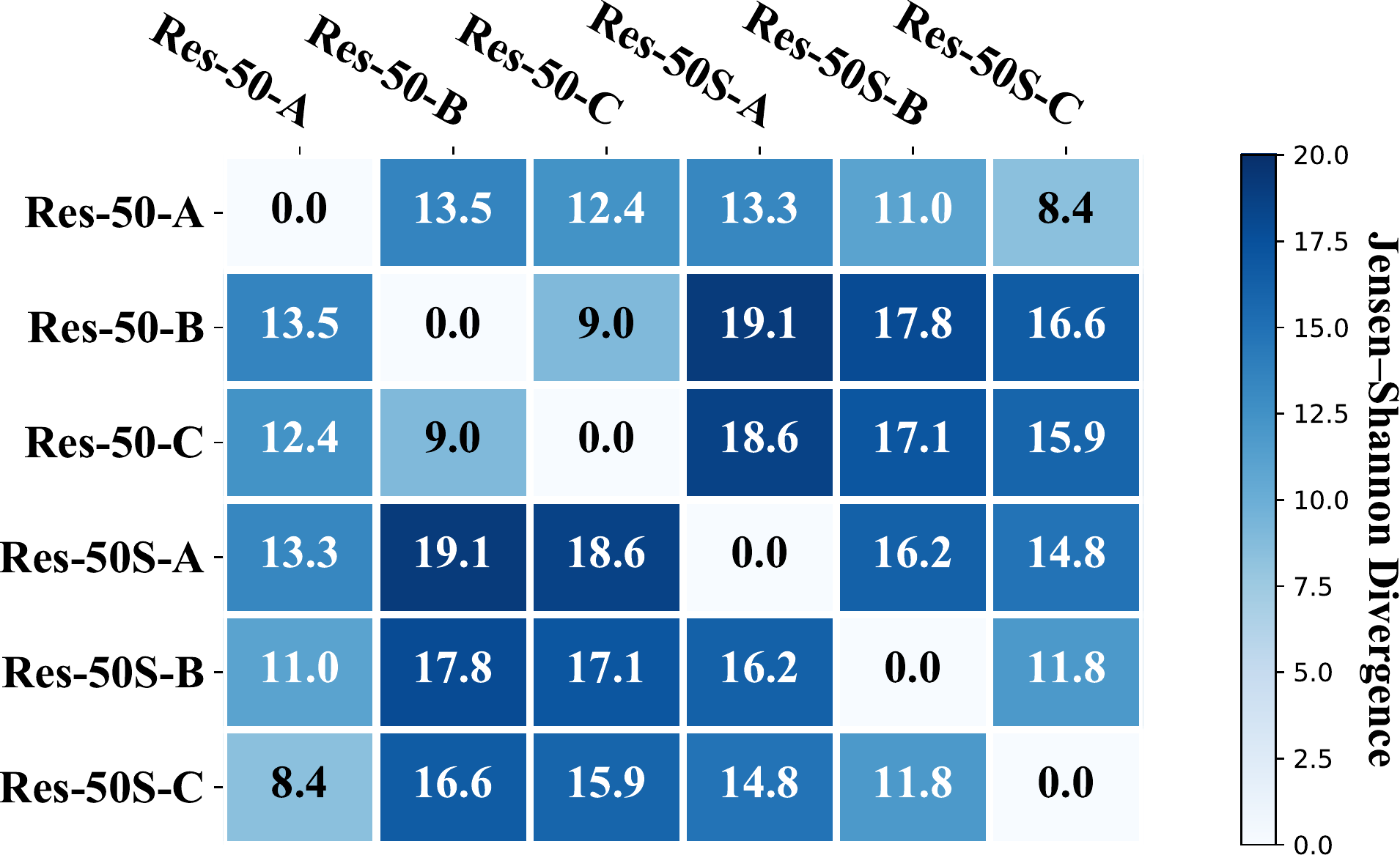}
\end{center}
% \vspace{-2ex}
\caption{The illustration of the mean diversity $(\times 10^{-2})$ of any pair of networks over the ILSVRC 2012 validation set.The higher value indicates larger diversity.}
\label{fig:jsd}
% \vspace{-3ex}
\end{figure}

\begin{table*}[tb]
\centering
\small
\begin{tabular}{|c|cccc||cc|cc|cc|cc|cc|cc||c|}
\hline
\multirow{2}{*}{Methods} & \multicolumn{4}{c||}{Settings} & \multicolumn{2}{c|}{Res-50} & \multicolumn{2}{c|}{Res-101} & \multicolumn{2}{c|}{Res-152} & \multicolumn{2}{c|}{IncRes-v2} & \multicolumn{2}{c|}{Inc-v3} & \multicolumn{2}{c||}{Inc-v4} & \multirow{2}{*}{CC}\\ 
\cline{2-17} 
& MT & \#S & \#L & \#I & I- & MI- & I- & MI- & I- & MI- & I- & MI- & I- & MI- & I- & MI- &\\ 
\hline
\hline
Exp.~S1 & $B$ & 1  & 1  &   1 & 16.3 & 29.4 & 17.8 & 31.3 & 16.7 & 29.6 &  8.3 & 20.0 &  5.3 & 13.7 &  7.3 & 18.4 & 1 \\ 
Exp.~S2 & $M$ & 1  & 1  &   1 &  8.4 & 17.4 &  6.1 & 19.9 &  6.4 & 17.9 &  5.7 & 15.2 &  1.7 &  5.6 &  1.9 &  7.2 & 1 \\ 
Exp.~S3 & $M$ & 1  & 10 &  10 & \textbf{23.4} & \textbf{39.4} & \textbf{23.7} & \textbf{40.1} & \textbf{21.1} & \textbf{38.0} & \textbf{11.2} & \textbf{26.8} &  \textbf{6.3} & \textbf{17.6} & \textbf{10.0} & \textbf{22.4} & 1 \\ \hline
Exp.~S4 & $M$ & 10 & 1  &  10 & 28.8 & 44.5 & 29.9 & 43.2 & 25.6 & 41.9 & 13.1 & 30.4 &  6.3 & 17.9 &  9.3 & 25.6 & 10 \\ 
Exp.~S5 & $M$ & 10 & 10 & 100 & \textbf{35.9} & \textbf{50.6} & \textbf{35.9} & \textbf{51.4} & \textbf{60.1} & \textbf{64.9} & \textbf{14.6} & \textbf{33.3} & \textbf{12.3} & \textbf{28.3} & \textbf{19.4} & \textbf{37.4} & 10 \\ 
\hline
\end{tabular}
% \vspace{-2ex}
\caption{The average black-box attack rate (\%) comparison of different methods over two iterative methods, ``I-" for I-FGSM and ``MI-" for MI-FGSM. MT denotes model type (either $B$ for the base model, or $M$ for ghost networks), \#I denotes the number of intrinsic models, and \#S (or \#L) denotes the number of models for standard (or longitudinal) ensemble in each iteration (branch). CC denotes the computational cost, which is a relative value and we set the CC of Exp.~S1 as $1$. We marked all highest attack success rate under the same CC in \textbf{boldface}.}
\label{table: SingleModelAttack}
% \vspace{-3ex}
\end{table*}

\begin{figure*}[tb]
\centering
\subfigure[]
{
\begin{minipage}[t]{0.235\textwidth}
\includegraphics[width = 1\textwidth]{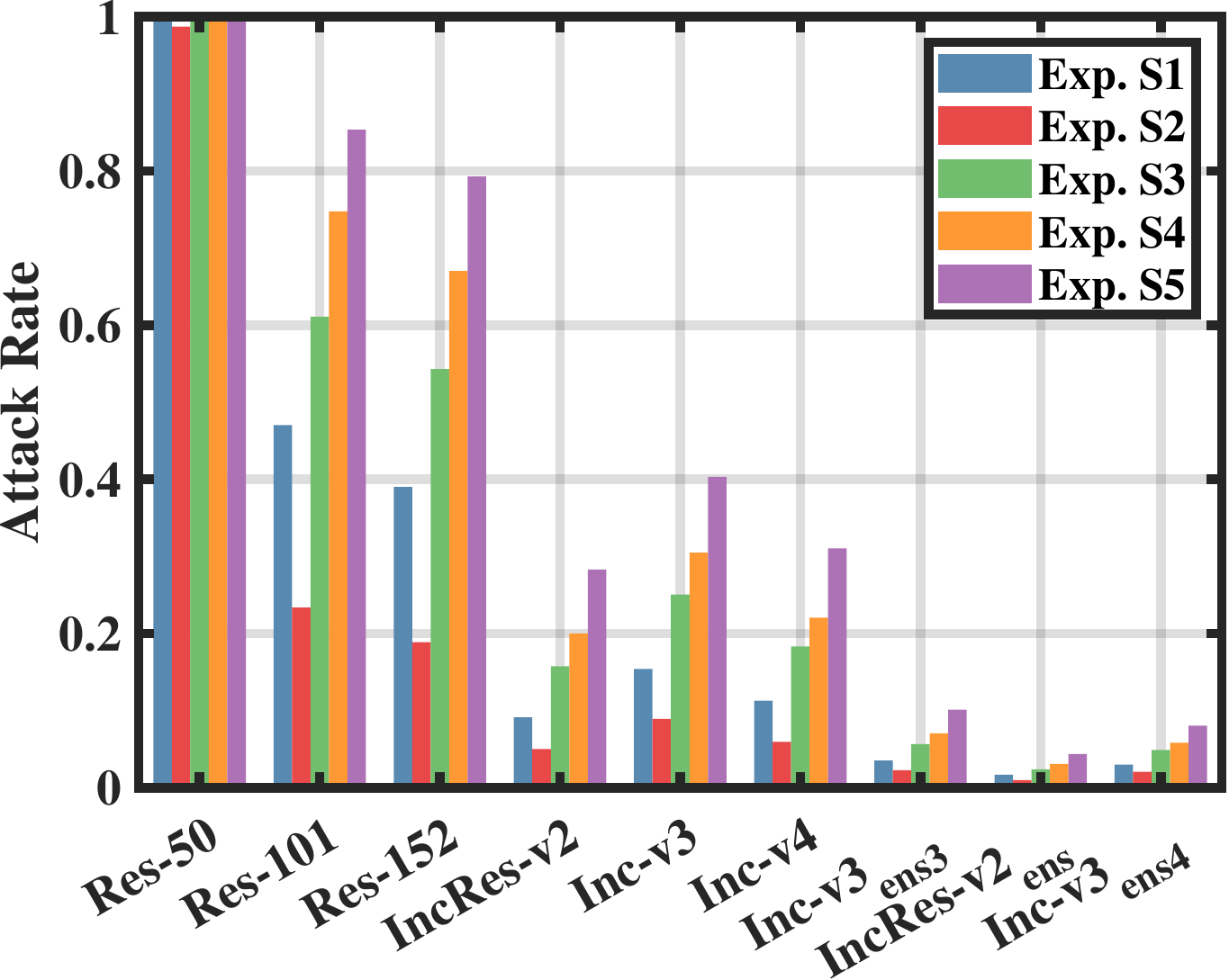}
\end{minipage}
\label{fig:IFGSM_Res50}
}
\subfigure[]
{
\begin{minipage}[t]{0.235\textwidth}
\includegraphics[width = 1\textwidth]{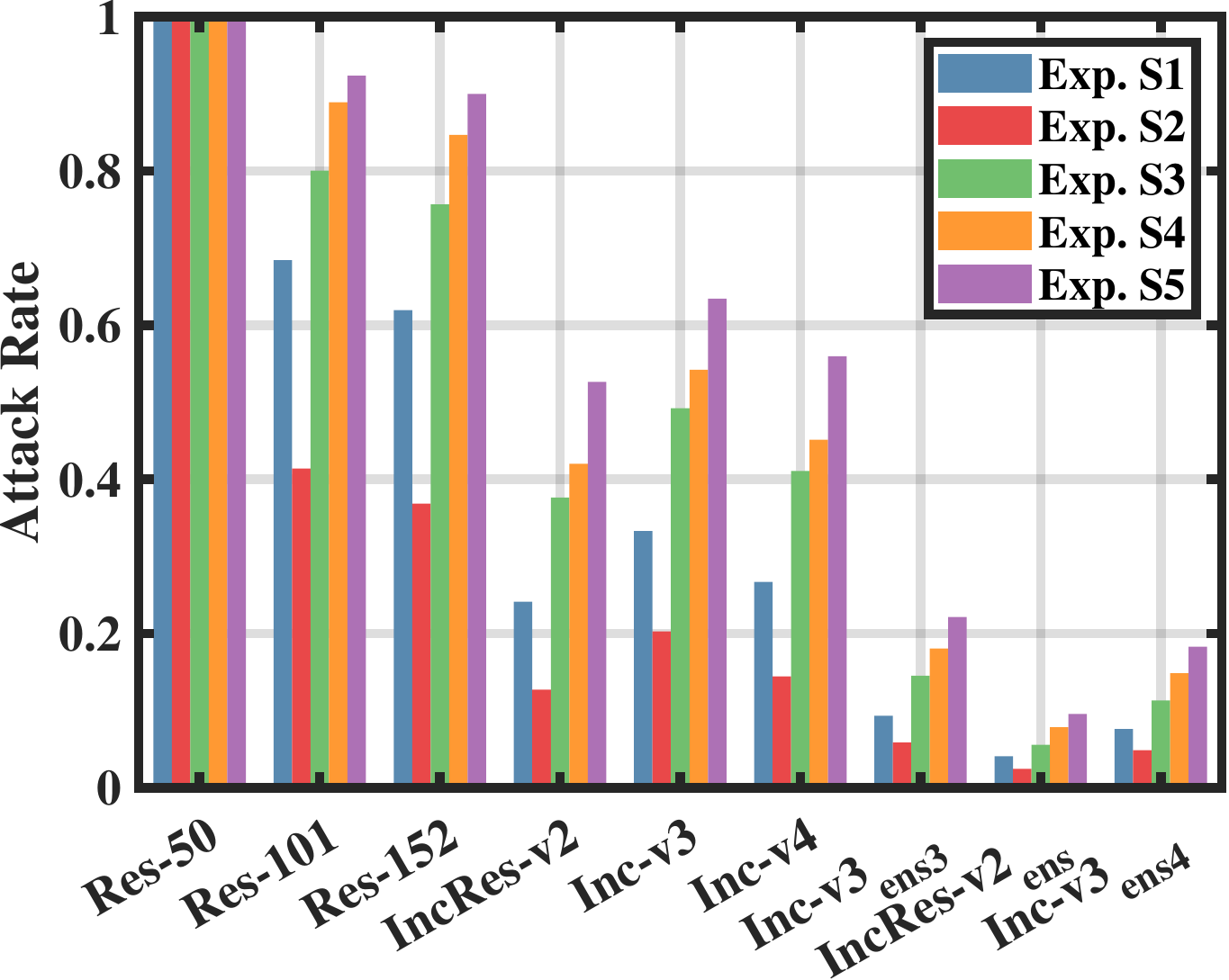}
\end{minipage}
\label{fig:MIFGSM_Res50}
}
\subfigure[]
{
\begin{minipage}[t]{0.235\textwidth}
\includegraphics[width = 1\textwidth]{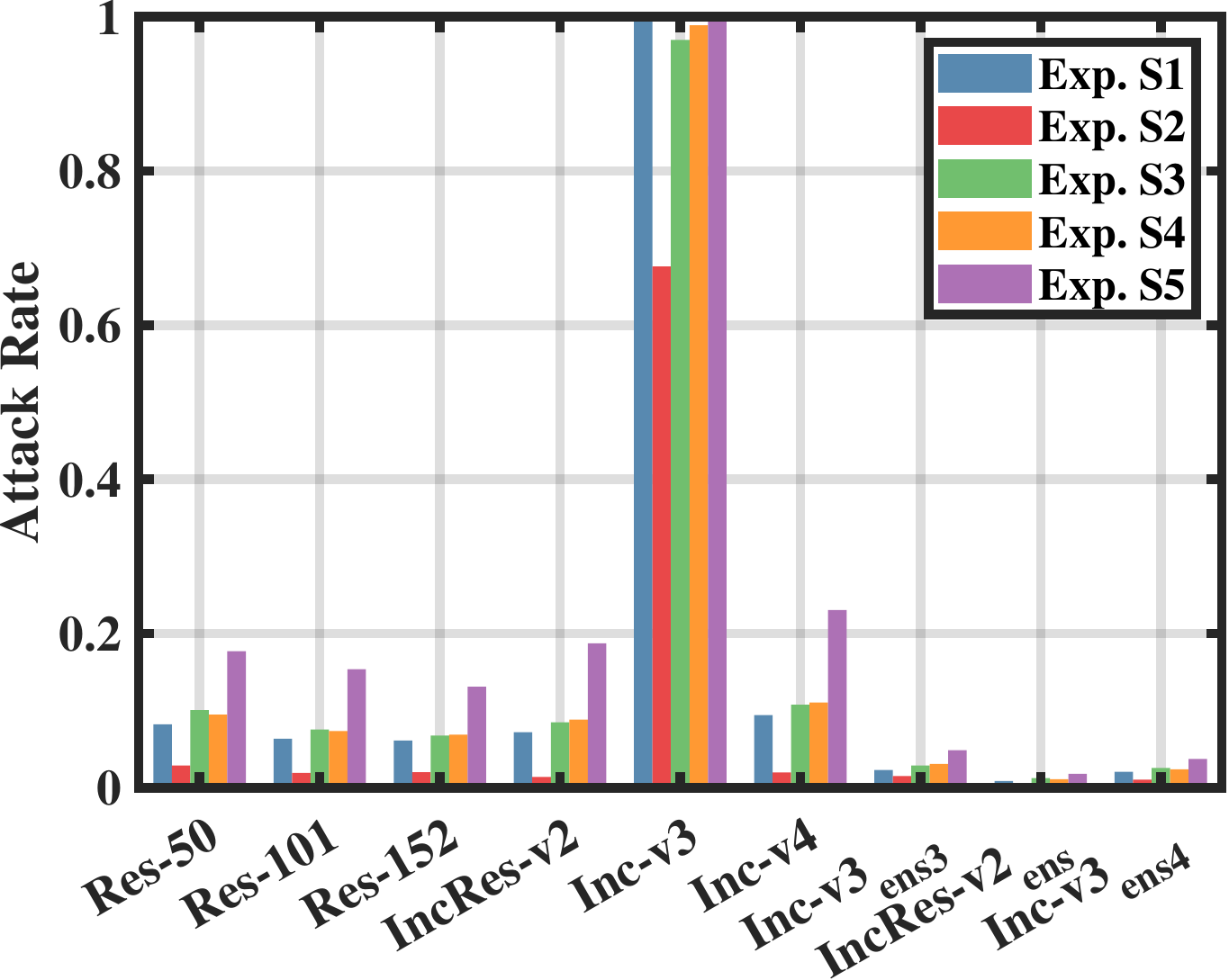}
\end{minipage}
\label{fig:IFGSM_Incv3}
}
\subfigure[]
{
\begin{minipage}[t]{0.235\textwidth}
\includegraphics[width = 1\textwidth]{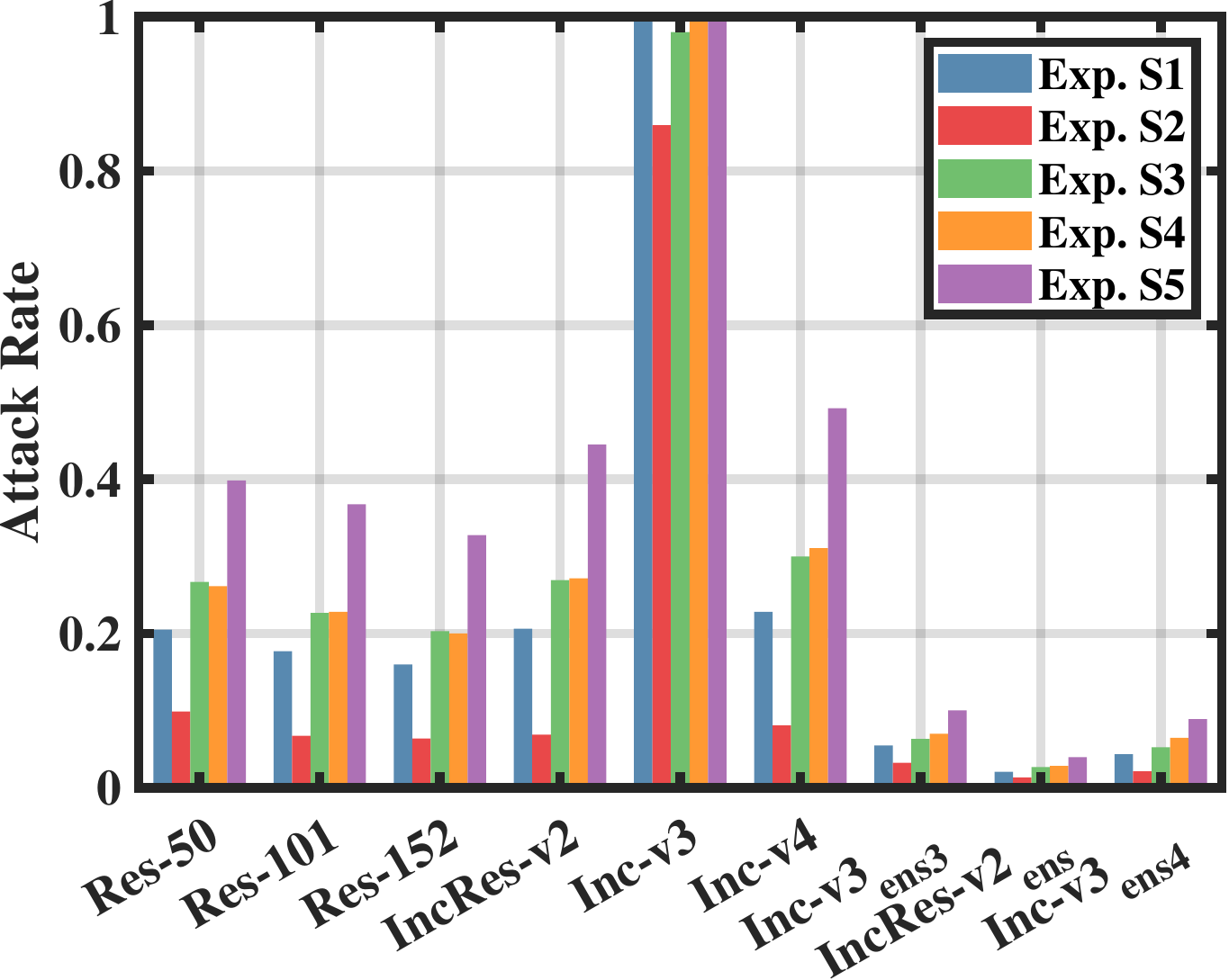}
\end{minipage}
\label{fig:MIFGSM_Incv3}
}
% \vspace{-3ex}
\caption{The attack rate (\%) comparison when attacking Res-50 (a)(b) and Inc-v3 (c)(d) with I-FGSM (a)(c) and MI-FGSM (b)(d).}
\label{fig:SingleAttackRate}
% \vspace{-3.5ex}
\end{figure*}

\begin{table}[tb]
\vspace{1ex}
\centering
\small
\begin{tabular}{|l|ccccc|cc|c|}
\hline
\multirow{2}{*}{Methods} & \multicolumn{5}{c|}{Settings} & \multicolumn{2}{c|}{Attack Rate} & \multirow{2}{*}{CC}  \\
\cline{2-8}
   & MT & \#B & \#S & \#L & \#I & I- & MI- &\\ 
\hline
\hline
Exp.~M1 & $B$ & 1 & 1 & 1 & 1  & 25.5 & 37.2 & 1 \\ 
Exp.~M2 & $B$ & 3 & 3 & 1 & 3  & 33.6 & 46.8 & 3 \\ 
Exp.~M3 & $M$ & 1 & 3 & 1 & 3  & 28.9 & 37.2 & 3 \\ 
Exp.~M4 & $M$ & 3 & 3 & 1 & 3  & 26.3 & 40.8 & 3 \\ 
Exp.~M5 & $M$ & 1 & 3 & 10& 30 & 38.3 & 52.5 & 3 \\ 
Exp.~M6 & $M$ & 3 & 3 & 10& 30 & \textbf{41.1} & \textbf{54.3} & 3 \\ 
\hline
\end{tabular}
% \vspace{-2ex}
\caption{The comparison of attack rate ($\%$) of multi-model attack. ``I-", ``MI-", MT, \#S, \#L, \#I and CC have the same meaning as in Table~\ref{table: SingleModelAttack}. \#B denotes the number of base models. We test on the $9$ networks described in the \textbf{Experimental Setup} section and report the average performances.}
\label{table: MultiModelAttack}
% \vspace{-3ex}
\end{table}

\section{Experiments}
In this section, we give a comprehensive experimental evaluation of the proposed Ghost Networks. In order to distinguish models trained from scratch and the ghost networks we generate, we call the former one the base network or base model in the rest of this paper. We release source code and provide additional experimental results in \url{https://github.com/LiYingwei/ghost-network}.

\subsection{Experimental Setup} \label{sec:setupExp}
\vspace{1ex}\noindent\textbf{Base Networks.}~$9$ base models are used in our experiments,  including $6$ normally trained models,~\ie,~Resnet v2-\{50, 101, 152\} (Res-\{50, 101, 152\})~\cite{he2016identity}, Inception v3 (Inc-v3)~\cite{szegedy2017inception}, Inception v4 (Inc-v4) and Inception Resnet v2 (IncRes-v2)~\cite{szegedy2016rethinking}, and $3$ adversarially-trained models~\cite{tramer2017ensemble}
,~\ie,~Inc-v3\textsubscript{ens3}, Inc-v3\textsubscript{ens4} and IncRes-v2\textsubscript{ens}.

\vspace{1ex}\noindent\textbf{Datasets.}~Because it is less meaningful to attack images that are originally misclassified, following \citet{xie2018improving}, we select $5000$ images from the ILSVRC 2012 validation set, which can be correctly classified by all the $9$ base models. 

\vspace{1ex}\noindent\textbf{Attacking Methods.}~We employ two iteration-based attack methods mentioned in the \textbf{Backgrounds} section to evaluate the adversarial robustness,~\ie,~I-FGSM and MI-FGSM.

\vspace{1ex}\noindent\textbf{Parameter Specification.}~If not specified otherwise, we follow the default settings in \citet{kurakin2016adversarial},~\ie,~step size $\alpha = 1$ and the total iteration number $N = \min(\epsilon+4, 1.25\epsilon)$. We set the maximum perturbation $\epsilon=8$ (the iteration number $N=10$ in this case). For the momentum term, the decay factor $\mu$ is set to be $1$ as in \citet{dong2017boosting}.

\subsection{Analysis of Ghost Networks} \label{sec:ghostNetworkExp}
In order to learn adversarial examples with good transferability, there are generally two requirements for the intrinsic models. First, each model should have a low test error. Second, different models should be diverse (\ie,~converge at different local minima). To show the generated ghost networks are qualified for adversarial attack, we experiment with the whole ILSVRC 2012 validation set.

\vspace{1ex}\noindent\textbf{Descriptive Capacity.}~In order to quantitatively measure the descriptive capacity of the generated ghost networks, we plot the relationship between the magnitude of erosion $\Lambda$ and top-1 classification accuracy. 

We apply dropout erosion to non-residual networks (Inc-v3 and Inc-v4) and skip connection erosion to residual networks (Res-50, Res-101, Res-152 and IncRes-v2). Fig.~\ref{fig:GhostAcc} (a) and (b) present the accuracy curves of the dropout erosion and skip connection erosion, respectively.

The classification accuracies of different models are negatively correlated to the magnitude of erosion $\Lambda$ as expected. By choosing the performance drop approximately equal to 10\% as a threshold, we can determine the value of $\Lambda$ individually for each network. Specifically, in our following experiments, $\Lambda$s are $0.006$, $0.012$, $0.22$, $0.16$, $0.12$ and $0.08$ for Inc-v3, Inc-v4, Res-50, Res-101, Res-152, and IncRes-v2 respectively unless otherwise specified. As emphasized throughout this paper, it is extremely cheap to generate a huge number of ghost networks that still preserve relative low error rates.

\vspace{1ex}\noindent\textbf{Model Diversity.}~To measure diversity, we use Res-50 as the backbone model. We denote the base Res-50 described in the \textbf{Experimental Setup} section as Res-50-A, and independently train two additional models with the same architecture, denoted by Res-50-B and Res-50-C. Meanwhile, we apply skip connection erosion to Res-50-A, then obtain three ghost networks denoted as Res-50S-A, Res-50S-B, and Res-50S-C, respectively.

We employ the Jensen-Shannon Divergence (JSD) as the evaluation metric for model diversity. Concretely, we compute the pairwise similarity of the output probability distribution (\ie,~the predictions after softmax layer) for each pair of networks as in~\citet{huang2017snapshot}. Given any image, let $X$ and $Y$ denote the softmax outputs of two networks, then
\begin{equation}
\text{JSD}(X||Y)=\frac{1}{2}D(X\|Z)+\frac{1}{2}D(Y\|Z),
\end{equation}
where $Z$ is the average of $X$ and $Y$,~\ie,~$Z=(X+Y)/2$. $D(\cdot)$ is the Kullback-Leibler divergence. 

In Fig.~\ref{fig:jsd}, we report the averaged JSD for all pairs of networks over the ILSVRC 2012 validation set. As can be drawn, the diversity among ghost networks is comparable or even more significant than independently trained networks.

Based on the analysis of descriptive capacity and model diversity, we can see that generated ghost networks can provide accurate yet diverse descriptions of the data manifold, which is beneficial to learn transferable adversarial examples as we will experimentally prove below.

\begin{table*}[tb]
\centering
\small
\begin{tabular}{|l|cccccc|ccc|}
\hline
\multirow{2}{*}{Methods} & \multicolumn{6}{c|}{Hold-out} & \multicolumn{3}{c|}{Ensemble} \\ 
\cline{2-10}
  & -Res-50 & -Res-101 & -Res-152 & -IncRes-v2 & -Inc-v3 & -Inc-v4 & Inc-v3\textsubscript{ens3} & Inc-v3\textsubscript{ens4} & IncRes-v2\textsubscript{ens}\\
\hline
\hline

I-FGSM         & 71.08 & 71.16 & 67.92 & 46.60 & 59.98 & 50.86 & 15.94 &  8.54 & 13.72 \\
I-FGSM + \textbf{ours}  & 80.22 & 79.80 & 77.02 & 60.20 & 73.18 & 67.84 & 25.80 & 13.56 & 21.42 \\
MI-FGSM        & 79.32 & 79.14 & 77.26 & 64.24 & 72.22 & 66.64 & 29.98 & 16.66 & 26.16 \\
MI-FGSM + \textbf{ours} & \textbf{87.14} & \textbf{86.14} & \textbf{84.64} & \textbf{74.18} & \textbf{82.06} & \textbf{79.18} & \textbf{39.56} & \textbf{21.24} & \textbf{32.68} \\
\hline

\end{tabular}
% \vspace{-2ex}
\caption{The attack rate (\%) comparison of multi-model attack. ``Ensemble" means attack all $6$ naturally-trained models. ``Hold-out" means attack $5$ out of $6$ models. The sign ``-" indicates the name of the hold-out model.}
\label{table: MultiModelAttack2}
% \vspace{-2ex}
\end{table*}

\begin{table*}[tb]
\centering
\small
\begin{tabular}{|p{3.73cm}|cccc|cccc|}
\hline
\multirow{2}{*}{Methods} & \multicolumn{4}{c|}{Black-box Attack}  & \multicolumn{4}{c|}{White-box Attack} \\
\cline{2-9}
& TsAIL & iyswim & Anil Thomas &  Average & Inc-v3\_adv & IncRes-v2\_ens & Inc-v3 & Average \\ 
\hline
\hline
No.1 Submission & 13.60 & 43.20 & 43.90 & 33.57 & 94.40 & 93.00 & \textbf{97.30} & 94.90 \\ 
% MI-FGSM & 14.14 & 47.40 & 46.90 & 36.15 & 98.60 & 95.66 & 97.20 & 97.15 \\ 
% MI-FGSM+\textbf{ours} & 14.90 & 55.50 & 54.28 & 41.56 & 97.70 & 96.42 & 97.64 & 97.25 \\ 
No.1 Submission+\textbf{ours} & \textbf{14.80} & \textbf{52.28} & \textbf{51.68} & \textbf{39.59} & \textbf{97.62} & \textbf{96.00} & 95.48 & \textbf{96.37} \\
\hline
\end{tabular}
% \vspace{-2ex}
\caption{The attack rate (\%) comparison in the NeurIPS 2017 Adversarial Challenge.}
\label{table:nips}
% \vspace{-3ex}
\end{table*}

\subsection{Single-model Attack} \label{sec:singleModelAttackExp}

Firstly, we evaluate the ghost networks in single-model attack, where attackers can only access one base model $B$ trained from scratch. To demonstrate the effectiveness of our method, we design five experimental comparisons. The setting, black-box attack success rate, and properties are shown in Table~\ref{table: SingleModelAttack}. The difference among five experiments is the type of model to attack, the number of models ensembled by standard ensemble~\cite{liu2016delving} in each iteration, and the number of models ensembled by longitudinal ensemble in each branch of the standard ensemble. For example, Exp.~S5 combines two ensemble methods, that is, we do a standard ensemble of $10$ models for each iteration of attack and a longitudinal ensemble of $10$ models. Therefore, in Exp.~S5, the intrinsic number of models is $100$.

% \vspace{1ex}\noindent\textbf{Exp.~S1}: We attack the base model $B$ by the two attack methods (I-FGSM or MI-FGSM) as two baselines.

% \vspace{1ex}\noindent\textbf{Exp.~S2}: We apply erosion to the base model and obtain one ghost network $M$. Then, an adversarial attack is conducted to $M$ to generate the adversarial examples.

% \vspace{1ex}\noindent\textbf{Exp.~S3}: We independently apply erosion $10$ times to get a pool of ghost networks $M=\{M_1,M_2,...,M_{10}\}$, then utilize the proposed longitudinal ensemble to efficiently fuse them during adversarial attack.

% \vspace{1ex}\noindent\textbf{Exp.~S4}: Similar to Exp.~S3, the only difference is that we use the standard ensemble method proposed in~\cite{liu2016delving} to fuse the $10$ ghost networks.

% \vspace{1ex}\noindent\textbf{Exp.~S5}: $100$ ghost networks are generated, which are fused in a $10\times 10$ manner, that is, we do a standard ensemble of $10$ models for each iteration of attack and a longitudinal ensemble of $10$ models.

We attack $6$ normally-trained networks and test on all the $9$ networks (include $3$ adversarially-trained networks). The attack rate is shown in Table~\ref{table: SingleModelAttack}. To save space, we report the average attack rate for black-box models. Individual cases are shown in Fig.~\ref{fig:SingleAttackRate}.

As can be drawn from Table~\ref{table: SingleModelAttack}, a single ghost network is worse than the base network (Exp.~S2 \emph{vs.} Exp.~S1), because the descriptive power of ghost networks is inferior to base networks. However, by leveraging the longitudinal ensemble, our work achieves a much higher attack rate in most settings (Exp.~S3 \emph{vs.} Exp.~S1).
%For example, when attacking Res-50 in black-box attack, Exp.~S3 outperforms Exp.~S1 by $7.1$ with I-FGSM and by $10.0$ with MI-FGSM. 
This observation firmly demonstrates the effectiveness of ghost networks in learning transferable adversarial examples. It should be mentioned that the computational cost of Exp.~S3 almost remains the same as Exp.~S1 for two reasons. First, the $10$ ghost networks used in Exp.~S3 are not trained but eroded from the base model and used on-the-fly. Second, multiple ghost networks are fused via the longitudinal ensemble, instead of the standard ensemble method in~\citet{liu2016delving}.

The proposed ghost networks can also be fused via the standard ensemble method, as shown in Exp.~S4. In this case, we achieve a higher attack rate at the sacrifice of computational efficiency. This observation inspires us to combine the standard ensemble and the longitudinal ensemble as shown in Exp.~S5. As we can see, Exp.~S5 consistently beats all the compared methods in all the black-box settings. Of course, Exp.~S5 is as computational expensive as Exp.~S4. However, the additional computational overhead stems from the standard ensemble rather than longitudinal ensemble. 

Note that in all the experiments presented in Table~\ref{table: SingleModelAttack}, we use only one individual base model. Even in the case of Exp.~S3, all the to-be-fused models are ghost networks. However, the generated ghost networks are never stored or trained, meaning no extra space complexity. Therefore, the benefit of ghost networks is obvious. Especially when comparing Exp.~S5 and Exp.~S1, ghost networks can achieve a substantial improvement in black-box attack.

Based on the experimental results above, we arrive at a similar conclusion as~\citet{liu2016delving}: the number of intrinsic models is essential to improve the transferability of adversarial examples. However, a different conclusion is that it is less necessary to train different models independently. Instead, ghost networks is a computationally cheap alternative enabling good performance. When the number of intrinsic models increases, the attack rate will increase. We will further exploit this phenomenon in multi-model attack.

In Fig.~\ref{fig:SingleAttackRate}, we select two base models,~\ie,~Res-50, and Inc-v3, to attack and present their performances when testing on all the $9$ base models. It is easy to observe the improvement of transferability by adopting ghost networks.

\subsection{Multi-model Attack} \label{sec:multiModelAttackExp}
We evaluate ghost networks in multi-model setting, where attackers have access to multiple base models.

\subsubsection{Same Architecture and Different Parameters}
We firstly evaluate a simple setting of multi-model attack, where base models share the same network architecture but have different weights. The same three Res-50 models as in the \textbf{Analysis of Ghost Networks} section are used. The settings of $6$ experiments are shown in Table~\ref{table: MultiModelAttack}. Besides a new parameter \#B (the number of trained-from-scratch models), others are the same as the single model attack setting. When \#B is $1$, we will use Res-50-A as the only one base model, and settings are the same as single-model attack. When \#B is $3$, \#S is always $3$, and each branch of the standard ensemble is assigned to a different base model. In Exp.~M4 and Exp.~M6, the ghost network(s) in each standard ensemble branch will be generated by the base model assigned to that branch.

% \vspace{1ex}\noindent\textbf{Exp.~M1}:~A standard ensemble of the base model $B_1$ for three times. This is simply equivalent to single-model attack, which can serve as a weak baseline.

% \vspace{1ex}\noindent\textbf{Exp.~M2}:~A standard ensemble of the base models $B_1$ and $B_2$ and $B_3$, which can serve as a strong baseline.

% \vspace{1ex}\noindent\textbf{Exp.~M3}:~A standard ensemble of $M_{1,1}$ and $M_{1,2}$ and $M_{1,3}$, which simply replaces the base model $B_1$ in Exp.~1 with three ghost networks associated to it.

% \vspace{1ex}\noindent\textbf{Exp.~M4}:~A standard ensemble of $M_{1,1}$ and $M_{2,1}$ and $M_{3,1}$, which replaces the base networks used in Exp.~2 with ghost networks, each one associated to a base model.

% \vspace{1ex}\noindent\textbf{Exp.~M5}:~$30$ ghost networks are generated upon the base model $B_1$. They are fused in a $3\times 10$ manner, that is, we do a standard ensemble of $3$ models for each iteration of attack, and a longitudinal ensemble of $10$ models in total.

% \vspace{1ex}\noindent\textbf{Exp.~M6}:~For each base model, we generate $10$ ghost networks. At the $j$-th iteration of attack, we do a standard ensemble of $\{M_{1,j},M_{2,j},M_{3,j}\}$, then do a longitudinal ensemble of $\{M_{i,1},M_{i,2},...,M_{i,10}\}$ for the $i$-th base model.

The adversarial examples generated by each method are used to test on all the $9$ models. We report the average attack rates in Table~\ref{table: MultiModelAttack}. It is easy to understand that Exp.~M2 performs better than Exp.~M1, Exp.~M3, and Exp.~M4 as it has three independently trained models. However, by comparing Exp.~M5 with Exp.~M2, we observe a significant improvement of attack rate. For example, By using MI-FGSM as the attack method, Exp.~M5 beats Exp.~M2 by $6.70$. 
Although Exp.~M5 only has $1$ base model and Exp.~M2 has $3$, Exp.~M5 actually fuses $30$ intrinsic models.  Such a result further supports our previous claim that the number of intrinsic models is essential, but it is less necessary to obtain them by training from scratch independently. Similarly, Exp.~M6 yields the best performance as it has $3$ independently trained models and $30$ intrinsic models.

\subsubsection{Different Architectures}
Besides the baseline comparison above, we then evaluate ghost networks in the multi-model setting following~\citet{liu2016delving}. We attack an ensemble of $5$ out of $6$ normally-trained models in this experiment, then test the hold-out network (black-box setting). We also attack an ensemble of $6$ normally-trained models and test on the $3$ adversarially-trained networks to evaluate the transferability of the generated adversarial examples in black-box attack.

The results are summarized in Table~\ref{table: MultiModelAttack2}, the performances in black-box attack are significantly improved. 
For example, when holding out Res-50, our method improves the performance of I-FGSM from $71.08$ to $80.22$, and that of MI-FGSM from $79.32$ to $87.14$. 
When testing on the three adversarially-trained networks, the improvement is more notable. These results further testify the ability of ghost networks to learn transferable adversarial examples.

\subsection{NeurIPS 2017 Adversarial Challenge} \label{sec: nipsExp}
Finally, we evaluate our method in a benchmark test of the NeurIPS 2017 Adversarial Challenge~\cite{kurakin2018adversarial}. For performance evaluation, we use the top-3 defense submissions (black-box models),~\ie,~TsAIL\footnote{https://github.com/lfz/Guided-Denoise}, iyswim\footnote{https://github.com/cihangxie/NIPS2017\_adv\_challenge\_defense} and Anil Thomas\footnote{https://github.com/anlthms/nips-2017/tree/master/mmd}, and three official baselines (white-box models),~\ie,~Inc-v3\textsubscript{adv}, IncRes-v2\textsubscript{ens} and Inc-v3. The test dataset contains $5000$ images with the same 1000-class labels as ImageNet~\cite{deng2009imagenet}.

Following the experimental setting of the No.1 attack submission~\cite{dong2017boosting}, we attack on an ensemble of Inc-v3, IncRes-v2, Inc-v4, Res-152, Inc-v3\textsubscript{ens3}, Inc-v3\textsubscript{ens4}, IncRes-v2\textsubscript{ens} and Inc-v3\textsubscript{adv}~\cite{kurakin2016scale}. The ensemble weights are set to $1$/$7.25$ equally for the first seven networks and $0.25$/$7.25$ for Inc-v3\textsubscript{adv}. The total iteration number is set to $10$, and the maximum perturbation $\epsilon$ is randomly selected from $\{4, 8, 12, 16\}$. The step size $\alpha = \epsilon/10$.
% Our settings are exactly the same with the No.1 attack submission~\cite{dong2017boosting}, but apply our method to each clean trained model.
The results are summarized in Table~\ref{table:nips}. Consistent with previous experiments, we observe that by applying ghost networks, the performance of the No. 1 submission can be significantly improved, especially with black-box attack.
For example, the average performance of black-box attack is changed from $33.57$ to $39.59$, an improvement of $6.02$. The most remarkable improvement is achieved when testing on iyswim, where ghost networks leads to an improvement of $9.08$. 
This suggests that our proposed method can generalize well to other defense mechanisms.

\section{Conclusion}
%This paper focuses on learning transferable adversarial examples for adversarial attacks. We propose, for the first time, to exploit network erosion to generate a kind of virtual models called ghost networks. Ghost networks, together with the coupled longitudinal ensemble strategy, requiring almost no additional time and space consumption, is an effective tool to improve existing methods in learning transferable adversarial examples. Extensive experiments have firmly demonstrated the efficacy of ghost networks. Meanwhile, one can potentially apply erosion to residual unit by other methods or \textbf{densely} erode other typical layers (\eg,~batch norm~\cite{ioffe2015batch} and relu~\cite{nair2010rectified}) \textbf{through} a neural network. We suppose these methods could improve the transferability as well, and leave these issues as future work.
This paper focuses on learning transferable adversarial examples for adversarial attacks. We propose, for the first time, to exploit network erosion to generate a kind of virtual models called ghost networks. Ghost networks, together with the coupled longitudinal ensemble strategy, is an effective and efficient tool to improve existing methods in learning transferable adversarial examples. Extensive experiments have firmly demonstrated the efficacy of ghost networks. Meanwhile, one can potentially apply erosion to residual unit by other methods or \textbf{densely} erode other typical layers (\eg,~batch norm~\cite{ioffe2015batch} and relu~\cite{nair2010rectified}) \textbf{through} a neural network. We suppose these methods could improve the transferability as well, and leave these issues as future work.

\vspace{1ex}\noindent\textbf{Acknowledgements}~This paper is supported by ONR award N00014-15-1-2356.

\fontsize{9.5pt}{10.5pt} \selectfont
\bibliography{ref.bib}
\bibliographystyle{aaai}
\end{document}